\documentclass{article}

 \usepackage[main,final]{neurips_2025}

\usepackage{natbib}                 
\setcitestyle{authoryear, open={(},close={)}}
\usepackage[utf8]{inputenc} 
\usepackage[T1]{fontenc}    
\usepackage{hyperref}       
\usepackage{url}            
\usepackage{booktabs}       
\usepackage{amsfonts}       
\usepackage{nicefrac}       
\usepackage{microtype}      
\usepackage{xcolor}         
\usepackage{graphicx}
\usepackage{algorithm}
\usepackage{algpseudocode}
\usepackage{float}
\usepackage{wrapfig}
\usepackage{listings}
\usepackage{inconsolata}
\usepackage{upquote} 
\usepackage{subcaption}
\usepackage{lipsum}
\usepackage{placeins}
\usepackage[most]{tcolorbox}

\definecolor{codebg}{HTML}{F7F7F7}
\definecolor{codeframe}{HTML}{CCCCCC}
\definecolor{kw}{HTML}{005CC5}      
\definecolor{str}{HTML}{22863A}     
\definecolor{com}{HTML}{6A737D}     
\definecolor{num}{HTML}{B31D28}
\lstdefinestyle{echopy}{
  language=Python,
  basicstyle=\ttfamily\small,        
  keywordstyle=\color{kw}\bfseries,
  stringstyle=\color{str},
  commentstyle=\color{com}\itshape,
  numberstyle=\tiny\color{com},
  showstringspaces=false,
  keepspaces=true,
  columns=fullflexible,
  breaklines=true,
  frame=single,
  rulecolor=\color{codeframe},
  backgroundcolor=\color{codebg},
  aboveskip=0pt,
  belowskip=0pt,
  xleftmargin=0.5em,
  xrightmargin=0.5em,
  tabsize=2,
}

\title{Sample-Efficient Online Learning in LM Agents \\via Hindsight Trajectory Rewriting}

\author{%
  Michael Y. Hu$^{1}$\thanks{Work done while interning at Microsoft. Code: \url{https://github.com/michahu/echo}} \quad
  Benjamin Van Durme$^{2}$ \quad 
  Jacob Andreas$^{2}$ \quad 
  Harsh Jhamtani$^{2}$ \\
  $^{1}$New York University \quad $^{2}$Microsoft \\
  \texttt{michael.hu@nyu.edu, hjhamtani@microsoft.com}
}

\begin{document}

\maketitle

\begin{abstract}
Language model (LM) agents deployed in novel environments often exhibit poor sample efficiency when learning from sequential interactions. This significantly hinders the usefulness of such agents in environments where interaction is costly (for example, when they interact with humans or reset physical systems). While a number of existing LM agent architectures incorporate various mechanisms for experience storage and reflection, they make limited use of LMs' abilities to directly generate or reason about full counterfactual trajectories.
We introduce ECHO (Experience Consolidation via Hindsight Optimization), a prompting framework that adapts hindsight experience replay from reinforcement learning for language model agents. ECHO generates optimized trajectories for alternative goals that could have been achieved during failed attempts, effectively creating synthetic positive examples from unsuccessful interactions. Our approach consists of two components: a hindsight rule that uses the language model itself to identify relevant subgoals and generate optimized trajectories, and an update rule that maintains compressed trajectory representations in memory. We evaluate ECHO on stateful versions of XMiniGrid, a text-based navigation and planning benchmark, and PeopleJoinQA, a collaborative information-gathering enterprise simulation. Across both domains, ECHO outperforms vanilla language agent baselines by up to 80\%; in XMiniGrid, it also outperforms a number of sophisticated agent architectures including Reflexion and AWM, demonstrating faster adaptation to novel environments through more effective utilization of past experiences.
\end{abstract}

\section{Introduction}

While language models (LMs) have demonstrated remarkable generalization across tasks, their performance often degrades in unfamiliar or interactive environments, especially when learning from limited experience \citep{ziems-etal-2024-large,kwa2025measuringaiabilitycomplete,liang2023holistic}. In such settings, sample efficiency becomes critical, particularly when interactions are costly (e.g., with humans or physical systems).  
For example, a conversational assistant deployed for the first time in a new organization likely does not know where to look for specific pieces of information, or the best means of communicating with specific people.
Thus, creating agents that can learn and adapt to their environments over time is of critical importance in improving their everyday usability.

Here we study the problem of building efficient mechanisms for \emph{online learning} in LM agents. We consider the setting where an LM agent receives queries one at a time in a streaming fashion. 
Existing LM agent frameworks typically approach this setting through reflection \citep{shinn2023reflexion, expel}, memory \citep{wang2025agent}, or experience replay mechanisms \citep{zheng2024synapse}, which allow agents to revisit past episodes and improve over time. However, these methods primarily focus on storing or synthesizing experiences, and thus fail to fully exploit the LM’s ability to reason about counterfactuals---what could have led to success in past failures. This gap suggests an opportunity to design LM agents that actively rewrite and optimize their past experiences, converting failures into synthetic successes that improve future decision-making. 

In this work, we introduce ECHO (Experience Consolidation via Hindsight Optimization), a framework that adapts hindsight experience replay (HER) to LM agents, enabling them to generate and learn from counterfactual trajectories for more sample-efficient learning.
Our approach builds on \textbf{hindsight experience replay} methods from the RL literature \citep[HER]{andrychowicz2017hindsight}. HER learns a goal-conditioned policy; during training, each attempt to reach a goal state $s$ that fails in an end state $s'$ is interpreted as a successful trajectory for reaching $s'$. For example, a trajectory in which an LM fails to slice an apple by
attempting to grab a knife, dropping it, then grasping an apple may still be interpreted as a demonstration of a successful grasp.
But HER and related methods are relatively limited in the set of trajectory modifications they can make---relabeling trajectories with goals, but not altering the structure demonstrated trajectories themselves.

ECHO is a significantly more expressive hindsight relabeling method for LMs. In ECHO, LMs can perform arbitrary re-writing of failed trajectories, including changing both their goals and their intermediate steps. In the running example, this procedure might not only relabel the failed slicing attempt as a successful grasp, but edit out the knife-grasping attempt that was relevant to the initial goal but not the relabeled one.

We test ECHO and various state-of-the-art agent architectures in stateful variants of a 2D GridWorld task (XMiniGrid, \cite{nikulin2023xlandminigrid}) and a question-answering task with multiple agents and tool calling (PeopleJoinQA, \cite{jhamtani-etal-2025-llm}). These environments require exploration to successfully solve all queries of the task. In XMiniGrid, the agent must explore to find various objects in different rooms, and in PeopleJoin, the agent has imprecise or incomplete information about which teammates have the information it needs to answer a question. We make these environments stateful by allowing agents to persist insights via a scratchpad memory. Next, we reset the environment to the same initial position or state and vary the queries or tasks posed to the agent. The agent can then infer information about the environment over time and improve its performance and efficiency.

On XMiniGrid-Stateful, ECHO outperforms the baseline reason-then-act (ReAct) LM agent \citep{yao2023react} by 80\% in average reward and the next best baseline by 42\%. On PeopleJoinQA-Stateful, ECHO still outperforms the standard ReAct agent on both accuracy and efficiency, while being slightly worse than the best baseline in accuracy by 4.6\% and tied in efficiency. We conclude that ECHO is a promising technique for improving the sample efficiency of language agents, especially in environments where rewards are sparse and the baseline language agent performs poorly.

\begin{figure*}[t!]
    \centering
    \includegraphics[width=0.8\linewidth,trim=0.1 3.4in 3.4in 1.2in]{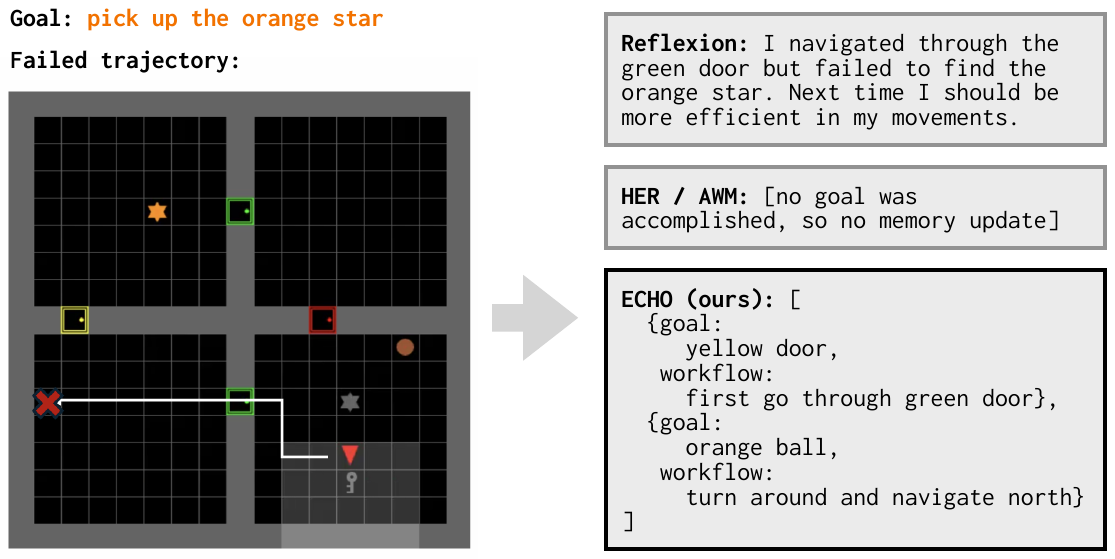}
    \caption{ECHO in the XMiniGrid environment. The agent fails in its first trajectory (left). Using this trajectory, ECHO identifies other objects the agent could have reached, and generates an optimized trajectory for these goals (right). In future iterations, the agent can then use these optimized trajectories to successfully complete unseen goals.}
    \label{fig:concept}
    \vspace{-1em}
\end{figure*}

\section{Related Work}

\paragraph{Language Model Agents:}
Language model (LM) agents are systems that use large language models as reasoning engines to interact with environments, make decisions, and execute actions over time \citep{yao2023react, schick2023toolformer}. These agents typically operate through a perception--action loop, where they observe their environment, reason about the current state, and generate actions \citep{wang2023survey}. Recent work has demonstrated LM agents' capabilities across diverse domains, from web navigation and tool use to multi-agent collaboration and code generation \citep{liu2024agentbench, chatdev, jhamtani-etal-2025-llm}. However, a key limitation of current LM agents is their reliance on static knowledge encoded during pre-training, making them less effective when deployed in novel environments that require exploration and adaptation \citep{zhou2024webarena}. This motivates the need for agents that can accumulate experience and improve their performance through interaction with their environment over time.

\paragraph{Offline Reasoning and Memory}
Following \citet{sumers2024cognitive}, we categorize memory systems for LM agents into two types: semantic and episodic. Semantic memory contains facts about the environment, and episodic memory stores past actions. In this work, we consider two baselines, Reflexion and Agent Workflow Memory (AWM), as exemplars of manipulating semantic and episodic memory \citep{shinn2023reflexion, wang2025agent}. Reflexion instructs the language model to reflect on the previous trajectory and propose areas of improvement; we consider these high-level notes about the environment to be part of semantic memory. AWM instructs the model to generate a summary workflow of the trajectory, provided the trajectory is successful; we consider this to be episodic memory. Building on this work, the current paper develops an improved mechanism for constructing and updating episodic memories.

\paragraph{Experience Replay}
One reason why off-policy RL algorithms can be more efficient than on-policy ones is that they can store and learn from informative trajectories that are low probability under the current policy, or even trajectories that were never observed at all. 
Such \emph{experience replay} techniques have proven especially valuable in situations with sparse rewards or limited data, as they extract maximal learning signal from a small number of positive examples \citep{PER, lu2023synthetic, zhang2025rlep}. In particular, hindsight experience replay (HER) further improves sample efficiency by relabeling past trajectories with alternative goals that were actually achieved during execution \citep{andrychowicz2017hindsight}. For instance, if an agent fails to reach a target location but successfully navigates to an intermediate point, HER treats this trajectory as a successful example for reaching that intermediate goal. \cite{codeit} apply HER to self-improve language models at writing code; our approach can be viewed as a generalization of HER in which not only goals, but arbitrary aspects of trajectories, can be edited in hindsight. \citet{bagel} also take inspiration from HER to induce natural language descriptions of agent trajectories; we go one step further and induce synthetic trajectories themselves.

\section{Approach}

\subsection{ECHO: Experience Consolidation via Hindsight Optimization}
\label{subsec:echo}

\begin{wrapfigure}{r}{0.5\textwidth}
\vspace{-2em}
\begin{lstlisting}[style=echopy, caption={ECHO pseudocode}, label={lst:echo}]
def ECHO(LM, trajectory, replay_buf={}):
    # hindsight rule
    summary = LM.summarize(trajectory)
    goals = LM.identify_goals(trajectory)
    for goal in goals:
        new_traj = LM.infer_traj(goal, trajectory)

        # update rule
        old_traj = replay_buf[goal]
        if old_traj and len(new_traj) < len(old_traj):
            replay_buf[goal] = new_traj
    return replay_buf
\end{lstlisting}
\vspace{-2em}
\end{wrapfigure}

We consider an online setting wherein an LM agent processes a sequence of queries from time $t=0$ to $T$ \emph{without} access to a ground-truth reward function or demonstrations.

Our key insight is that LMs have sufficient world knowledge to propose general edits to a trajectory, in addition to simply relabeling the trajectory for a particular goal. We take inspiration from HER and design a prompting framework that allows language agents to modify their past experiences. We call this framework Experience Consolidation via Hindsight Optimization, or ECHO. 
The basic idea behind ECHO is to take an existing trajectory and identify not just what goals that trajectory achieves, but all goals for which a successful trajectory can be \emph{synthesized} given the initial rollout.

ECHO contains two parts: a hindsight rule and an update rule (Listing \ref{lst:echo}). During application of the hindsight rule, the LM first proposes goals that it can \emph{infer} how to accomplish from a given trajectory.
If no goals are proposed, then ECHO does nothing. Next, the LM generates an optimized trajectory or description from the goal and the original trajectory. The optimized trajectory or description is given in natural language; see Figure \ref{fig:concept} or \S \ref{sec:examples} for examples.

In the update rule, for each entity, we compare its newly generated descriptor to the descriptor's predecessor and save the shorter workflow. Our motivation here is related to Kolmogorov complexity, or minimum description length (see \citet{grunwald2007minimum} for an overview); we want the replay buffer to contain the shortest possible description for achieving the goal.

ECHO runs the risk of appending a very short trajectory description early in the sequence of interactions, after which future trajectories will be ineffective. In our experience, this is very rare, because the LM has the option to abstain, or not propose any goals. As such, the goal-trajectory pairs that are added to the replay buffer are valid or near-valid trajectories. Nevertheless, a more precise update rule is a fruitful area for future work (\S \ref{sec:future-work}).

\subsection{Baselines}

In Listings \ref{lst:reflexion} and \ref{lst:awm}, we cast Reflexion \citep{shinn2023reflexion} and AWM \citep{wang2025agent} into our hindsight-update framework as well. The important parts of these algorithms lie in the hindsight rule, and the update rule simply appends the hindsights edits to memory.

\begin{figure}[!ht]
    \centering 
    
    \begin{minipage}[t]{0.48\textwidth}
        \begin{lstlisting}[style=echopy, caption={Reflexion pseudocode.}, label={lst:reflexion}]
def reflexion(LM, trajectory, replay_buf=[]):
    # hindsight rule
    reflection = LM.reflect(trajectory)

    # update rule
    replay_buf.append(reflection)
    return replay_buf
        \end{lstlisting}
    \end{minipage}
    \hfill 
    \begin{minipage}[t]{0.48\textwidth}
        \begin{lstlisting}[style=echopy, caption={AWM pseudocode.}, label={lst:awm}]
def AWM(LM, trajectory, replay_buf=[]):
    # hindsight rule
    if LM.is_successful(trajectory):
        workflow = LM.infer_workflow(trajectory)

    # update rule
    replay_buf.append(workflow)
    return replay_buf
        \end{lstlisting}
    \end{minipage}
\end{figure}

\subsection{Qualitative Example}
\label{sec:examples}

In this section, we show responses from Reflexion, AWM, and ECHO after the same trajectory, in which an agent failed to pick up a grey key.

\begin{tcolorbox}[
    colback=black!5!white, 
    colframe=black!75!white, 
    title=Reflexion,
    fonttitle=\bfseries
]
In this episode, I successfully navigated towards the grey key but failed to pick it up despite multiple attempts. I need to ensure that I correctly execute the 'pick up' action when I am adjacent to the target item. Additionally, I should be more efficient in my movements and avoid unnecessary turns and steps. In future episodes, I will focus on better spatial awareness and precise action execution to achieve my goals more effectively.
\end{tcolorbox}

\begin{tcolorbox}[colback=black!5!white, colframe=black!75!white, title=AWM, fonttitle=\bfseries]
\{
  "goal": "Pick up grey key.",
  "workflow": ""
\}
\end{tcolorbox}

\begin{tcolorbox}[
    colback=black!5!white, 
    colframe=black!75!white, 
    title=ECHO,
    fonttitle=\bfseries
]
\{
  "goal": "Pick up the grey star",
  "workflow": "Step 1: Navigate north from the starting location. Step 2: Move towards the grey star located to the northeast. Step 3: Pick up the grey star."
\}
\end{tcolorbox}

Reflexion notes that the agent failed, but its feedback is generic; empirically, this kind of feedback does not change the model's performance within XMiniGrid-Stateful (\S \ref{subsec:xminigrid-results}). Since the trajectory failed, AWM correctly declines to generate a workflow. Conversely, ECHO successfully notices that the agent also observed the grey star while unsuccessfully navigating towards the grey key, producing a correct optimized trajectory explaining how to pick up the grey star.

\section{Results}
\label{sec:results}

To test our algorithm, we create stateful versions of two environments where exploration is crucial: XMiniGrid \citep{nikulin2023xlandminigrid} and PeopleJoinQA \citep{jhamtani-etal-2025-llm}. These corresponding versions, XMiniGrid-Stateful and PeopleJoinQA-Stateful, can be used as benchmarks for efficient online learning in language agents; we release these environments at \url{https://github.com/michahu/echo}. We ran all experiments using GPT-4o \citep{hurst2024gpt}; see Appendix \ref{app:hyperparams} for hyperparameters.

\subsection{Evaluation}  We refer to the full sequence of states, actions, and rewards encountered as an episode, and a timestep as a single state-action-reward tuple within an episode. The offline algorithms we consider here operate between episodes.

Our evaluation metrics are \textbf{final average reward} (or accuracy) and \textbf{cumulative average reward}. The cumulative average reward at episode $\tau$ is the average of all rewards received up to that episode:
$$\text{Cumulative Average Reward at } \tau = \frac{1}{\tau+1} \sum_{t=0}^{\tau} R_t$$
where $R_t$ is the reward achieved in episode $t$. We use this metric to compare agents' sample efficiency, as sample-efficient agents will rapidly increase the cumulative average reward. 
To normalize for problem difficulty, we report rewards as improvements over a baseline ReAct agent (\citealp{yao2023react}). Thus, the best method will be the one that maximizes both the final average reward and the rate at which it improves upon the ReAct agent.

\subsection{XMiniGrid-Stateful}
\label{subsec:xminigrid-results}

XMiniGrid is a procedurally-generated GridWorld, where an agent navigates and perform tasks in a partially-observable 2D grid environment. XMiniGrid takes inspiration from XLand, a suite of procedurally-generated, partially-observable 3D games. To create XMiniGrid-Stateful, we prompt the agent to achieve one randomly sampled goal in the same environment per episode and reset the agent and other objects in the environment to the same starting locations between episodes. Thus, the agent can learn the starting locations of unseen objects over time. 

In total, we test the language models on 10 unique, procedurally generated environments. Each environment has 4 objects distributed across 4 rooms. Partial observability makes the task challenging, akin to picking up objects in a dimly-lit house. For each environment, we sequentially ask the model to pick up a randomly sampled object 16 times, allowing duplicate queries. For each query, we give the model up to 64 steps to achieve it. Thus, the maximum number of queries required to run XMiniGrid-Stateful is $10 \times 16 \times 64 = 10,240$.

To make XMiniGrid compatible with language models, we convert its 2D observation space to an egocentric text description, which reads something like ``You are two steps from a wall. You see a red door two steps to the right.'' Since even navigation in this partially observed environment is challenging for LMs, we restrict the randomly sampled goals to ``pick up'' goals. XMiniGrid-Stateful's evaluation metric is mean reward, or success rate of picking up the object, over the 64 steps.

\begin{figure}[h!] 
    \centering 
    
    \begin{subfigure}{0.48\textwidth}
        \centering
        \includegraphics[width=\linewidth]{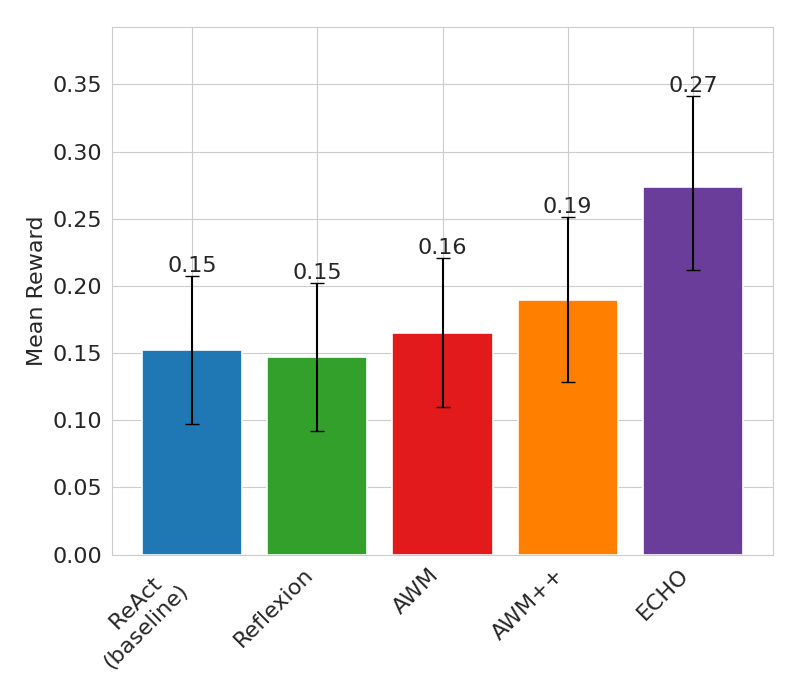}
    \end{subfigure}
    \hfill 
    \begin{subfigure}{0.48\textwidth}
        \centering
        \includegraphics[width=\linewidth]{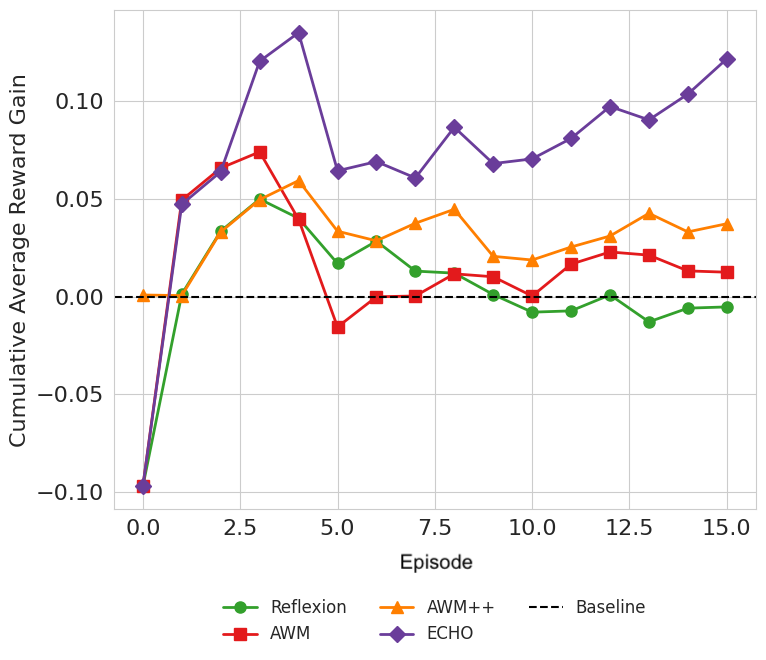}
    \end{subfigure}
    
    \caption{Results on the XMiniGrid-Stateful benchmark. Left: ECHO achieves the highest mean reward. Right: ECHO's cumulative reward is higher than the baseline ReACT agent's after 3 interactions, indicating that ECHO improves compared to a static baseline over time.}
    \label{fig:xminigrid}
\end{figure}

ECHO strictly outperforms all other methods on XMiniGrid (Figure \ref{fig:xminigrid}) In addition to AWM, we created a baseline called AWM++ which replaces AWM's update rule with our own (keeping the shorter workflow in memory when a goal collision occurs). Performance with this update rule improves slightly, but not enough to recapture the entire improvement from ECHO.

\paragraph{Trajectory Validity Analysis:} While ECHO leverages the LM’s world knowledge to synthesize counterfactual trajectories, these may not always be executable under true environment dynamics. To assess this limitation, we evaluated the validity of the hindsight-imputed workflows generated by ECHO. For each synthesized trajectory–goal pair, we then provided the full hindsight-imputed workflow to the LM agent as part of the system prompt. Across 40 sampled examples from XMiniGrid, the agent successfully reached these imputed goals 85\% of the time (34 / 40). 4 of the remaining 6 failures arose from agent deviations from instructions during execution and 2 were due to infeasible steps in the synthesized workflows. This indicates that the counterfactual workflows generated by ECHO in XMiniGrid are largely correct and lead the agent to successful solutions.

\subsection{PeopleJoinQA-Stateful}

PeopleJoinQA is a question-answering environment where an agent must synthesize (join) information collected from simulated people to answer a question. To contact people and retrieve information, the agent must also use various tools, such as organization directory and document search functions. PeopleJoinQA is also partially observable because the agent does not know ahead of time which people possess the information it is seeking. 

To create PeopleJoinQA-Stateful, we simply fix the organization, or the set of simulated people and knowledge, and ask the agent all the questions written for that organization. Between queries, the agent can then reflect on how information is distributed in the organization and improve. We chose 5 organizations within PeopleJoinQA, each with different numbers of people and possible queries. In total, there are 248 queries across the 5 organizations, and each query takes on average 7.98 messages between organization members to resolve, requiring a total of $1,980$ queries to run PeopleJoinQA-Stateful. (In our tests, XMiniGrid-Stateful is still slightly faster to run, due to the agent's observations being shorter).

\begin{figure}[h!] 
    \centering 
    \begin{subfigure}{0.48\textwidth}
        \centering
        \includegraphics[width=\linewidth]{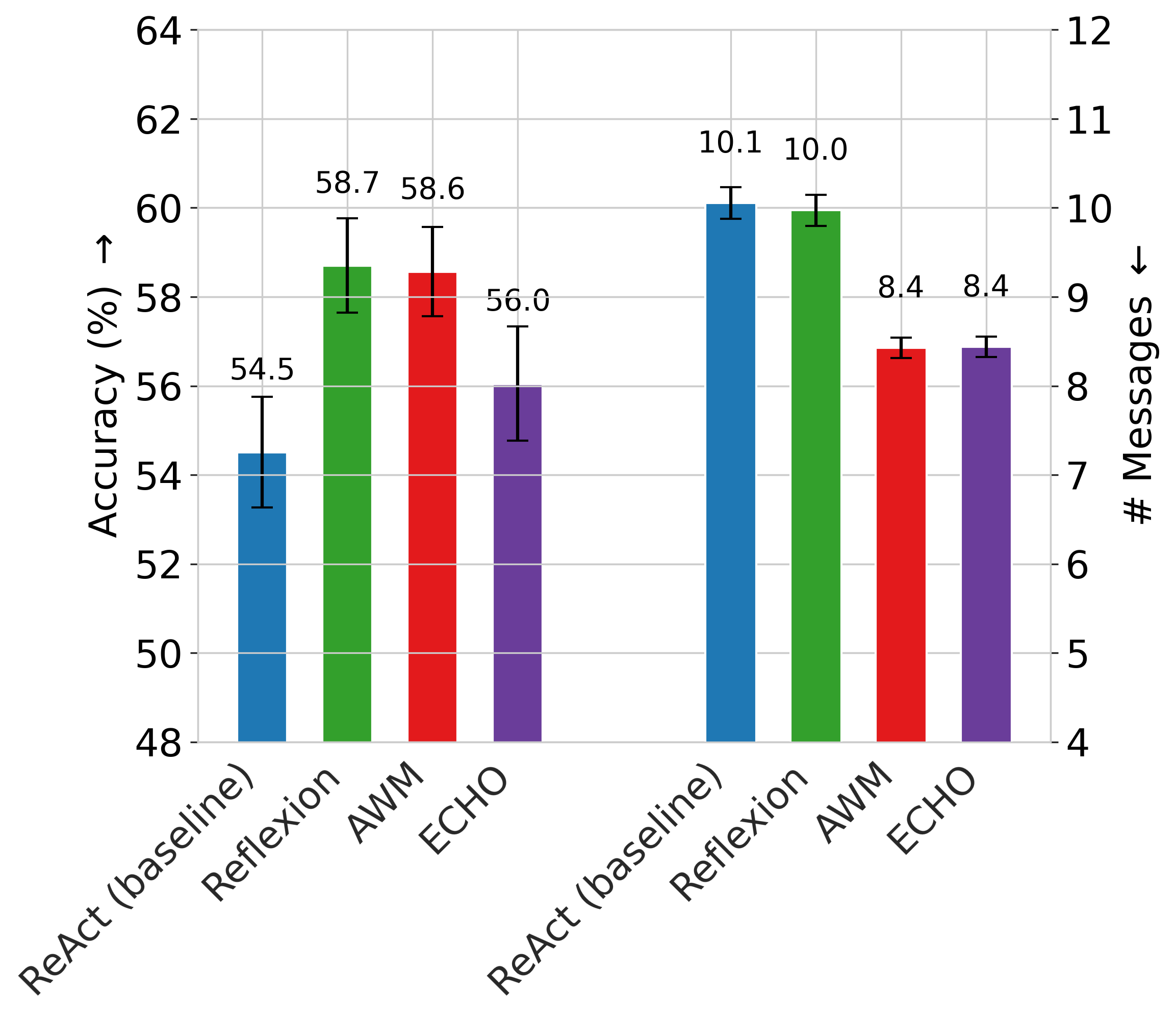}
    \end{subfigure}
    \hfill 
    \begin{subfigure}{0.48\textwidth}
        \centering
        \includegraphics[width=\linewidth]{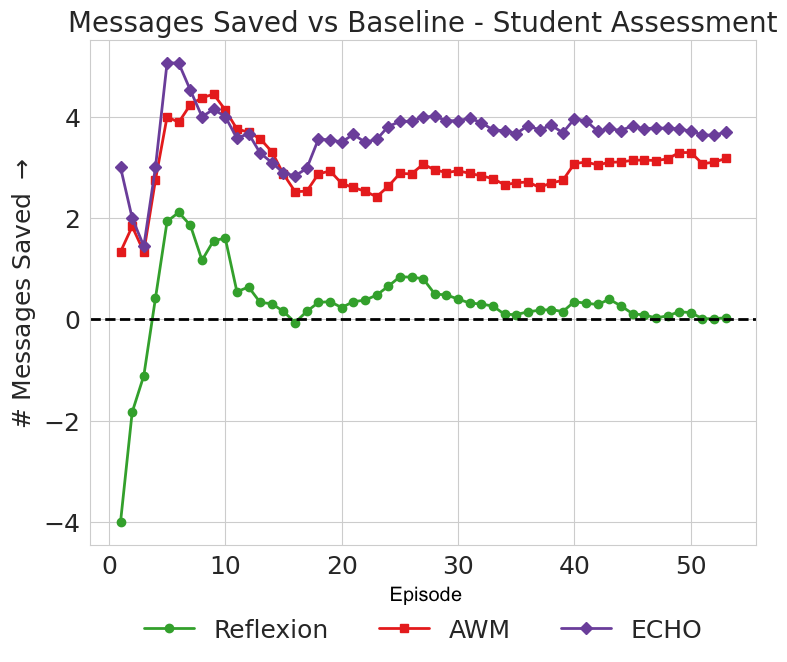}
    \end{subfigure}
    
    \caption{Results on the PeopleJoinQA-Stateful benchmark. Left: While Reflexion achieves slightly higher accuracy, ECHO and AWM are more efficient, completing the task in 1.6 fewer messages on average. Right: we plot the running average reward gain above ReAct. On average, ECHO outperforms the ReAct after the first query.}
    \label{fig:peoplejoinqa}
\end{figure}

In Figure \ref{fig:peoplejoinqa}, the most accurate method for PeopleJoinQA-Stateful is Reflexion, which is notable because in Reflexion the model provides feedback to itself generically. Thus, methods that manipulate episodic memory like AWM or ECHO may not be always be helpful in improving accuracy. However, AWM and ECHO both improve the efficiency of the agent's interactions, decreasing the average number of messages sent between the agent and other people in the organization by around 1.6 messages. 

In fact, the most accurate or efficient method varies depending on the PeopleJoin organization, as shown in Figure \ref{fig:peoplejoinqa-breakdown} in the appendix; no method strictly dominates all other methods. For example, Figure \ref{fig:peoplejoinqa} (right) shows that ECHO becomes the most efficient method after around 15 total queries in the organization answering questions about department stores. Thus, although the offline methods we consider in this work clearly outperform the baseline in the aggregate, understanding how to improve the robustness of these offline methods for all settings is an important area for future work.

\section{Discussion}

\subsection{Language Models as Incomplete World Models}

After a single trajectory within an environment, the LM often will not have sufficient information to infer a full internal world model of its environment. However, we can use the LM's general knowledge to infer local improvements to its previous trajectory. Hindsight optimization uses the LM's pretrained world knowledge to fill in gaps and propose reasonable counterfactual information, even when the agent's direct experience is limited. By sidestepping the need for a complete world model, ECHO is particularly effective in partially observable environments where building world models would be infeasible.

\subsection{Connections between RL and Prompting}

ECHO continues the connection identified in ReACT \citep{yao2023react} and Reflexion \citep{shinn2023reflexion} between reinforcement learning techniques and prompting strategies for language model agents. Tratitional RL methods like hindsight experience replay rely on numerical rewards and states, whereas language models can operate over experiences that can be described and modified through natural language. This enables a more flexible form of experience replay where the agent actively edits and improves past experiences based on its linguistic and commonsense understanding of the task. ECHO also bridges experience-based learning and symbolic reasoning, leveraging language models as both world models and policy generators.

\subsection{Limitations and Future Work}
\label{sec:future-work}

In this work, we primarily considered natural language representations of semantic and episodic information. Recent work shows that code-like representations can be even more effective \citep{wang2025asi}, so future work should explore how ECHO's performance changes when outputting programmatic trajectory representations. Furthermore, our update heuristic of accepting shorter trajectories for the same task can likely be improved in favor of a more sophisticated that combines new and old information while maintaining the same bias towards compression.
Future work could also explore augmenting ECHO with retrieval-based mechanisms that draw from a memory bank  (\cite{zheng2024synapse,moghe2024interpreting}), enabling more targeted reuse of relevant experiences.

\section{Conclusion}

In this work, we introduced ECHO, a framework that improves LM agent sample efficiency by adapting hindsight experience replay from off-policy RL. Our results demonstrate that agents can effectively learn from past experiences by using themselves as incomplete world models to edit and optimize previous trajectories. To evaluate ECHO and other agents in stateful environments, we introduced two new adaptations of existing benchmarks: XMiniGrid-Stateful and PeopleJoinQA-Stateful, both of which require exploration and multi-step reasoning. By using the language model to propose and refine its own experiences, ECHO provides a path towards more sample-efficient and adaptable LM agents, particularly in partially observable environments with sparse feedback.

\bibliography{neurips_2025}

@inproceedings{
yao2023react,
title={ReAct: Synergizing Reasoning and Acting in Language Models},
author={Shunyu Yao and Jeffrey Zhao and Dian Yu and Nan Du and Izhak Shafran and Karthik R Narasimhan and Yuan Cao},
booktitle={The Eleventh International Conference on Learning Representations },
year={2023},
url={https://openreview.net/forum?id=WE_vluYUL-X}
}

@book{grunwald2007minimum,
  title={The {M}inimum {D}escription {L}ength {P}rinciple},
  author={Gr{\"u}nwald, Peter D},
  year={2007},
  publisher={MIT press}
}

@inproceedings{
schick2023toolformer,
title={Toolformer: Language Models Can Teach Themselves to Use Tools},
author={Timo Schick and Jane Dwivedi-Yu and Roberto Dessi and Roberta Raileanu and Maria Lomeli and Eric Hambro and Luke Zettlemoyer and Nicola Cancedda and Thomas Scialom},
booktitle={Thirty-seventh Conference on Neural Information Processing Systems},
year={2023},
url={https://openreview.net/forum?id=Yacmpz84TH}
}

@misc{wang2023survey,
      title={A Survey on Large Language Model based Autonomous Agents}, 
      author={Lei Wang and Chen Ma and Xueyang Feng and Zeyu Zhang and Hao Yang and Jingsen Zhang and Zhiyuan Chen and Jiakai Tang and Xu Chen and Yankai Lin and Wayne Xin Zhao and Zhewei Wei and Ji-Rong Wen},
      year={2023},
      eprint={2308.11432},
      archivePrefix={arXiv},
      primaryClass={cs.AI}
}

@article{chatdev,
    title = {ChatDev: Communicative Agents for Software Development},
    author = {Chen Qian and Wei Liu and Hongzhang Liu and Nuo Chen and Yufan Dang and Jiahao Li and Cheng Yang and Weize Chen and Yusheng Su and Xin Cong and Juyuan Xu and Dahai Li and Zhiyuan Liu and Maosong Sun},
    journal = {arXiv preprint arXiv:2307.07924},
    url = {https://arxiv.org/abs/2307.07924},
    year = {2023}
}

@article{hurst2024gpt,
  title={{GPT}-4o system card},
  author={Hurst, Aaron and Lerer, Adam and Goucher, Adam P and Perelman, Adam and Ramesh, Aditya and Clark, Aidan and Ostrow, AJ and Welihinda, Akila and Hayes, Alan and Radford, Alec and others},
  journal={arXiv preprint arXiv:2410.21276},
  year={2024}
}

@misc{kwa2025measuringaiabilitycomplete,
      title={Measuring AI Ability to Complete Long Tasks}, 
      author={Thomas Kwa and Ben West and Joel Becker and Amy Deng and Katharyn Garcia and Max Hasin and Sami Jawhar and Megan Kinniment and Nate Rush and Sydney Von Arx and Ryan Bloom and Thomas Broadley and Haoxing Du and Brian Goodrich and Nikola Jurkovic and Luke Harold Miles and Seraphina Nix and Tao Lin and Neev Parikh and David Rein and Lucas Jun Koba Sato and Hjalmar Wijk and Daniel M. Ziegler and Elizabeth Barnes and Lawrence Chan},
      year={2025},
      eprint={2503.14499},
      archivePrefix={arXiv},
      primaryClass={cs.AI},
      url={https://arxiv.org/abs/2503.14499}, 
}

@inproceedings{codeit,
  author={Natasha Butt and Blazej Manczak and Auke J. Wiggers and Corrado Rainone and David W. Zhang and Michaël Defferrard and Taco Cohen},
  title={CodeIt: Self-Improving Language Models with Prioritized Hindsight Replay},
  year={2024},
  cdate={1704067200000},
  url={https://openreview.net/forum?id=SXVn5IFsrs},
  booktitle={ICML},
}

@article{
liang2023holistic,
title={Holistic Evaluation of Language Models},
author={Percy Liang and Rishi Bommasani and Tony Lee and Dimitris Tsipras and Dilara Soylu and Michihiro Yasunaga and Yian Zhang and Deepak Narayanan and Yuhuai Wu and Ananya Kumar and Benjamin Newman and Binhang Yuan and Bobby Yan and Ce Zhang and Christian Cosgrove and Christopher D Manning and Christopher Re and Diana Acosta-Navas and Drew A. Hudson and Eric Zelikman and Esin Durmus and Faisal Ladhak and Frieda Rong and Hongyu Ren and Huaxiu Yao and Jue WANG and Keshav Santhanam and Laurel Orr and Lucia Zheng and Mert Yuksekgonul and Mirac Suzgun and Nathan Kim and Neel Guha and Niladri S. Chatterji and Omar Khattab and Peter Henderson and Qian Huang and Ryan Andrew Chi and Sang Michael Xie and Shibani Santurkar and Surya Ganguli and Tatsunori Hashimoto and Thomas Icard and Tianyi Zhang and Vishrav Chaudhary and William Wang and Xuechen Li and Yifan Mai and Yuhui Zhang and Yuta Koreeda},
journal={Transactions on Machine Learning Research},
issn={2835-8856},
year={2023},
url={https://openreview.net/forum?id=iO4LZibEqW},
note={Featured Certification, Expert Certification, Outstanding Certification}
}

@inproceedings{bagel,
author = {Murty, Shikhar and Manning, Christopher D. and Shaw, Peter and Joshi, Mandar and Lee, Kenton},
title = {BAGEL: bootstrapping agents by guiding exploration with language},
year = {2024},
publisher = {JMLR.org},
booktitle = {Proceedings of the 41st International Conference on Machine Learning},
articleno = {1498},
numpages = {17},
location = {Vienna, Austria},
series = {ICML'24}
}

@article{ziems-etal-2024-large,
    title = "Can Large Language Models Transform Computational Social Science?",
    author = "Ziems, Caleb  and
      Held, William  and
      Shaikh, Omar  and
      Chen, Jiaao  and
      Zhang, Zhehao  and
      Yang, Diyi",
    journal = "Computational Linguistics",
    volume = "50",
    number = "1",
    month = mar,
    year = "2024",
    address = "Cambridge, MA",
    publisher = "MIT Press",
    url = "https://aclanthology.org/2024.cl-1.8/",
    doi = "10.1162/coli_a_00502",
    pages = "237--291"
}

@inproceedings{
    nikulin2023xlandminigrid,
    title={{XL}and-MiniGrid: Scalable Meta-Reinforcement Learning Environments in {JAX}},
    author={Alexander Nikulin and Vladislav Kurenkov and Ilya Zisman and Viacheslav Sinii and Artem Agarkov and Sergey Kolesnikov},
    booktitle={Intrinsically-Motivated and Open-Ended Learning Workshop, NeurIPS2023},
    year={2023},
    url={https://openreview.net/forum?id=xALDC4aHGz}
}

@inproceedings{
lu2023synthetic,
title={Synthetic Experience Replay},
author={Cong Lu and Philip J. Ball and Yee Whye Teh and Jack Parker-Holder},
booktitle={Thirty-seventh Conference on Neural Information Processing Systems},
year={2023},
url={https://openreview.net/forum?id=6jNQ1AY1Uf}
}

@misc{zhang2025rlep,
      title={RLEP: Reinforcement Learning with Experience Replay for LLM Reasoning}, 
      author={Hongzhi Zhang and Jia Fu and Jingyuan Zhang and Kai Fu and Qi Wang and Fuzheng Zhang and Guorui Zhou},
      year={2025},
      eprint={2507.07451},
      archivePrefix={arXiv},
      primaryClass={cs.CL},
      url={https://arxiv.org/abs/2507.07451}, 
}

@inproceedings{andrychowicz2017hindsight,
 author = {Andrychowicz, Marcin and Wolski, Filip and Ray, Alex and Schneider, Jonas and Fong, Rachel and Welinder, Peter and McGrew, Bob and Tobin, Josh and Pieter Abbeel, OpenAI and Zaremba, Wojciech},
 booktitle = {Advances in Neural Information Processing Systems},
 editor = {I. Guyon and U. Von Luxburg and S. Bengio and H. Wallach and R. Fergus and S. Vishwanathan and R. Garnett},
 pages = {},
 publisher = {Curran Associates, Inc.},
 title = {Hindsight Experience Replay},
 url = {https://proceedings.neurips.cc/paper_files/paper/2017/file/453fadbd8a1a3af50a9df4df899537b5-Paper.pdf},
 volume = {30},
 year = {2017}
}

@inproceedings{PER,
  author       = {Tom Schaul and
                  John Quan and
                  Ioannis Antonoglou and
                  David Silver},
  editor       = {Yoshua Bengio and
                  Yann LeCun},
  title        = {Prioritized Experience Replay},
  booktitle    = {4th International Conference on Learning Representations, {ICLR} 2016,
                  San Juan, Puerto Rico, May 2-4, 2016, Conference Track Proceedings},
  year         = {2016},
  url          = {http://arxiv.org/abs/1511.05952},
  bibsource    = {dblp computer science bibliography, https://dblp.org}
}

@inproceedings{jhamtani-etal-2025-llm,
    title = "{LLM} Agents for Coordinating Multi-User Information Gathering",
    author = "Jhamtani, Harsh  and
      Andreas, Jacob  and
      Van Durme, Benjamin",
    editor = "Che, Wanxiang  and
      Nabende, Joyce  and
      Shutova, Ekaterina  and
      Pilehvar, Mohammad Taher",
    booktitle = "Findings of the Association for Computational Linguistics: ACL 2025",
    month = jul,
    year = "2025",
    address = "Vienna, Austria",
    publisher = "Association for Computational Linguistics",
    url = "https://aclanthology.org/2025.findings-acl.916/",
    doi = "10.18653/v1/2025.findings-acl.916",
    pages = "17800--17826",
    ISBN = "979-8-89176-256-5"
}

@inproceedings{
zhou2024webarena,
title={WebArena: A Realistic Web Environment for Building Autonomous Agents},
author={Shuyan Zhou and Frank F. Xu and Hao Zhu and Xuhui Zhou and Robert Lo and Abishek Sridhar and Xianyi Cheng and Tianyue Ou and Yonatan Bisk and Daniel Fried and Uri Alon and Graham Neubig},
booktitle={The Twelfth International Conference on Learning Representations},
year={2024},
url={https://openreview.net/forum?id=oKn9c6ytLx}
}

@inproceedings{
liu2024agentbench,
title={AgentBench: Evaluating {LLM}s as Agents},
author={Xiao Liu and Hao Yu and Hanchen Zhang and Yifan Xu and Xuanyu Lei and Hanyu Lai and Yu Gu and Hangliang Ding and Kaiwen Men and Kejuan Yang and Shudan Zhang and Xiang Deng and Aohan Zeng and Zhengxiao Du and Chenhui Zhang and Sheng Shen and Tianjun Zhang and Yu Su and Huan Sun and Minlie Huang and Yuxiao Dong and Jie Tang},
booktitle={The Twelfth International Conference on Learning Representations},
year={2024},
url={https://openreview.net/forum?id=zAdUB0aCTQ}
}

@inproceedings{
shinn2023reflexion,
title={Reflexion: language agents with verbal reinforcement learning},
author={Noah Shinn and Federico Cassano and Ashwin Gopinath and Karthik R Narasimhan and Shunyu Yao},
booktitle={Thirty-seventh Conference on Neural Information Processing Systems},
year={2023},
url={https://openreview.net/forum?id=vAElhFcKW6}
}

@inproceedings{
wang2025agent,
title={Agent Workflow Memory},
author={Zora Zhiruo Wang and Jiayuan Mao and Daniel Fried and Graham Neubig},
booktitle={Forty-second International Conference on Machine Learning},
year={2025},
url={https://openreview.net/forum?id=NTAhi2JEEE}
}

@article{
sumers2024cognitive,
title={Cognitive Architectures for Language Agents},
author={Theodore Sumers and Shunyu Yao and Karthik Narasimhan and Thomas Griffiths},
journal={Transactions on Machine Learning Research},
issn={2835-8856},
year={2024},
url={https://openreview.net/forum?id=1i6ZCvflQJ},
note={Survey Certification}
}

@inproceedings{
zheng2024synapse,
title={Synapse: Trajectory-as-Exemplar Prompting with Memory for Computer Control},
author={Longtao Zheng and Rundong Wang and Xinrun Wang and Bo An},
booktitle={The Twelfth International Conference on Learning Representations},
year={2024},
url={https://openreview.net/forum?id=Pc8AU1aF5e}
}

@misc{wang2025asi,
      title={Inducing Programmatic Skills for Agentic Tasks}, 
      author={Zora Zhiruo Wang and Apurva Gandhi and Graham Neubig and Daniel Fried},
      year={2025},
      eprint={2504.06821},
      archivePrefix={arXiv},
      primaryClass={cs.CL},
      url={https://arxiv.org/abs/2504.06821}, 
}

@inproceedings{expel,
author = {Zhao, Andrew and Huang, Daniel and Xu, Quentin and Lin, Matthieu and Liu, Yong-Jin and Huang, Gao},
title = {ExpeL: LLM agents are experiential learners},
year = {2024},
isbn = {978-1-57735-887-9},
publisher = {AAAI Press},
url = {https://doi.org/10.1609/aaai.v38i17.29936},
doi = {10.1609/aaai.v38i17.29936},
booktitle = {Proceedings of the Thirty-Eighth AAAI Conference on Artificial Intelligence and Thirty-Sixth Conference on Innovative Applications of Artificial Intelligence and Fourteenth Symposium on Educational Advances in Artificial Intelligence},
articleno = {2188},
numpages = {11},
series = {AAAI'24/IAAI'24/EAAI'24}
}

@inproceedings{moghe2024interpreting,
  title={Interpreting User Requests in the Context of Natural Language Standing Instructions},
  author={Moghe, Nikita and Xia, Patrick and Andreas, Jacob and Eisner, Jason and Van Durme, Benjamin and Jhamtani, Harsh},
  booktitle={Findings of the Association for Computational Linguistics: NAACL 2024},
  pages={4043--4060},
  year={2024},
  url={https://aclanthology.org/2024.findings-naacl.255.pdf}
}


\appendix

\section{Prompts}
\label{app:prompts}

All of the offline proactive reasoning methods we study (Reflexion, AWM, ECHO, and their variants) are paired with a ReAct agent that performs actions.

\subsection{XMiniGrid-Stateful}

\begin{tcolorbox}[
    colback=black!5!white, 
    colframe=black!75!white, 
    title=ReACT,
    fonttitle=\bfseries
]
You are an agent in a 2D gridworld. At each step you will receive a list of valid and invalid actions. Choose a valid action by its index. Complete the goal in \#HORIZON\# steps. \\

You will be prompted at each turn to first reason about your plan and then choose actions. \\

Reply concisely with following JSON format:
\{"thought": X, "choice": Y\} where X is your reasoning and Y is the index of the desired choice. Ensure Y is a parseable integer!
\end{tcolorbox}

To help the agent understand which actions are valid or invalid, at each step we provide the dynamic lists \texttt{valid\_actions=\{1: ``go forward", ...\}, invalid\_actions=\{4:``pick up'', ...\}}.

\begin{tcolorbox}[
    colback=black!5!white, 
    colframe=black!75!white, 
    title=Reflexion,
    fonttitle=\bfseries
]
You are an agent in a 2D text-based environment. Reflect on your performance in the following episode and write some concise notes on how you can improve your performance in the next episodes. Reply with the following JSON format: \{"reflection": X\}
where X is your reflection. Ensure X is a parsable string!
\end{tcolorbox}

\begin{tcolorbox}[
    colback=black!5!white, 
    colframe=black!75!white, 
    title=AWM,
    fonttitle=\bfseries
]
You are an agent in a 2D text-based environment. If the agent succeeds at accomplishing the given goal in the episode, convert the actions done in the following episode into abstract summary workflow. Discuss in high-level terms the steps a future agent should take to reach the goal. Include potential obstacles and landmarks in your workflow explanation. \\

Reply with the following JSON format: \{"goal": "X", "workflow": Y\} where X is the achieved goal and Y is your summary workflow. Ensure X and Y are parsable strings! \\

If the agent did not achieve the goal, then make Y an empty string.
\end{tcolorbox}

ECHO has 3 LM calls, which we name according to our pseudocode, reproduced below:
\vspace{1em}
\begin{lstlisting}[style=echopy]
def ECHO(LM, trajectory, replay_buf={}):
    # hindsight rule
    summary = LM.summarize(trajectory)
    goals = LM.identify_goals(trajectory)
    for goal in goals:
        new_traj = LM.infer_traj(goal, trajectory)

        # update rule
        old_traj = replay_buf[goal]
        if old_traj and len(new_traj) < len(old_traj):
            replay_buf[goal] = new_traj
    return replay_buf
\end{lstlisting}

\begin{tcolorbox}[
    colback=black!5!white, 
    colframe=black!75!white, 
    title=ECHO: LM.summarize,
    fonttitle=\bfseries
]
You are an expert at analyzing agent behavior in 2D text-based environments. Create a concise, high-level summary of the agent's trajectory. \\

\#\# Instructions: \\

**What to Include:**\\
- Group low-level actions into high-level behaviors (e.g., "explored northern corridor" not individual moves) \\
- **All** objects discovered \\
- Completed objectives \\

**What to Exclude:** \\
- Individual movement steps, redundant actions, minor environmental details \\

**Format:** Chronological entries representing distinct phases or achievements \\

\#\# Output Format: \\
\{
  "0": "Agent spawned in [location] and observed [key objects/features]",
  "1": "Agent navigated to [destination] and discovered [important findings]",
  "2": "Agent interacted with [object/entity] resulting in [outcome]",
  ...
\}
\end{tcolorbox}

\begin{tcolorbox}[
    colback=black!5!white, 
    colframe=black!75!white, 
    title=ECHO: LM.identify\_goals,
    fonttitle=\bfseries
]
You are an expert at analyzing 2D text-based environments to identify potential agent objectives. Given a trajectory summary, extract all possible goals an agent could pursue. The agent's goal will always be to pick up a specific object. \\

\#\# Task: \\
Identify all objects that could serve as pickup targets based on the environmental context shown in the summary. \\

\#\# Requirements: \\
- **Extract specific objects** mentioned in the trajectory
- Avoid locations or non-portable objects

\#\# Output Format: \\
\{
  "possible\_goals": [
    "Pick up the [object1]",
    "Pick up the [object2]",
    ...
  ]
\}
\end{tcolorbox}

\begin{tcolorbox}[
    colback=black!5!white, 
    colframe=black!75!white, 
    title=ECHO: LM.infer\_traj,
    fonttitle=\bfseries
]
You are an expert at creating action plans for agents in 2D text-based environments. Given a specific goal and a summary of a previous agent's actions, create a high-level workflow to achieve the goal. \\

\#\# Task:
Design an abstract workflow for accomplishing the given goal using the environmental features from the trajectory summary. \\

\#\# Requirements:

- **Environment-specific actions only**: reference actual locations, objects, or features from the summary  

- Use high-level abstractions (e.g., "navigate to the blue door")  

- **Avoid generic phrases** like "move toward goal" or "find the object"  

- Start from the agent's known starting location  

- Focus on strategic phases, not individual actions \\

\#\# Output Format:
\{
  "goal": "[provided goal]",
  "workflow": "Step 1: [specific environment action]. Step 2: [specific environment action]. Step 3: [etc.]"
\}
\end{tcolorbox}

\subsection{PeopleJoinQA-Stateful}

For the decision policy, we used the prompts from \citet{jhamtani-etal-2025-llm}, available here: \url{https://github.com/microsoft/peoplejoin}.

\begin{tcolorbox}[
    colback=black!5!white, 
    colframe=black!75!white, 
    title=Reflexion,
    fonttitle=\bfseries
]
You are a helpful and clever teacher of language agents. You have access to a prior interaction between a language agent and other agents in an organization, as well as your own reflection about the organization. Using the prior interaction and reflection, write a better  reflection that will help a future language agent perform better in this organization. \\

Structure your reflection in the following json format: \{'reflection': reflection\}, where reflection is a string. The reflection should be concise and focused on giving instructions to future agents in this organization.
\end{tcolorbox}

\begin{tcolorbox}[
    colback=black!5!white, 
    colframe=black!75!white, 
    title=AWM,
    fonttitle=\bfseries
]
You are a helpful and clever teacher of language agents. Attached below is a prior interaction between a language agent and other agents in an organization. If you deem the interaction to successfully and accurately answer the initial question, return a summary of the interaction so future agents can easily reference what to do in similar situations. The summary should contain the query, a summary of events, and the final answer. \\

If the interaction was successful, return a json \{'successful': true, 'summary': summary\}, where summary is a string. If the interaction was not successful, return a json 
\{'successful': false, 'summary': ''\}.
\end{tcolorbox}

For PeopleJoin, we found summarizing unnecessary. Furthermore, we found that it worked better to ask the model for one optimized trajectory instead of several. Because the number of timesteps in some PeopleJoin environments are long, we observed that adding many trajectories to memory at a time leads to the context length expanding significantly, even with our update rule.

\begin{tcolorbox}[
    colback=black!5!white, 
    colframe=black!75!white, 
    title=ECHO,
    fonttitle=\bfseries
]
You are a helpful and clever teacher of language agents. Given a trajectory, write a simplified counterfactual workflow and final answer. If the trajectory is already efficient, you can simply summarize the events. If the correct final answer is unclear, then do not generate a workflow or final answer. \\

The counterfactual trajectory should include: \\
- the query \\
- a workflow for solving the query \\
- the final answer \\

Return a json \{`query': query, `workflow': workflow, `final\_answer': final\_answer\}. If either the correct workflow or final answer are unclear, then you should not generate a workflow or final answer. To abstain, return empty strings for `workflow' and `final\_answer': \{`query': query, `workflow': `', `final\_answer': `'.
\end{tcolorbox}

\pagebreak

\section{Hyperparameters}

The GPT-4o hyperparameters we used for the agent itself versus the offline proactive reasoning are slightly different. Below, we have labeled the agent as ``ReACT'' and the offline reasoning as ``Offline.''

\begin{table}[h!]
\centering
\caption{XMiniGrid-Stateful, ReACT}
\begin{tabular}{@{}ll@{}}
\toprule
\textbf{Hyperparameter} & \textbf{Value} \\
\midrule
Temperature & 0 \\
Max New Tokens & 4000 \\
API Version & 05-13 \\
\bottomrule
\end{tabular}
\end{table}

\begin{table}[h!]
\centering
\caption{XMiniGrid-Stateful, Offline}
\begin{tabular}{@{}ll@{}}
\toprule
\textbf{Hyperparameter} & \textbf{Value} \\
\midrule
Temperature & 0 \\
Max New Tokens & 4000 \\
API Version & 05-13 \\
\bottomrule
\end{tabular}
\end{table}

\begin{table}[h!]
\centering
\caption{PeopleJoin, ReACT}
\begin{tabular}{@{}ll@{}}
\toprule
\textbf{Hyperparameter} & \textbf{Value} \\
\midrule
Temperature & 0 \\
Max New Tokens & 3800 \\
API Version & 11-20 \\
\bottomrule
\end{tabular}
\end{table}

\begin{table}[h!]
\centering
\caption{PeopleJoin, Offline}
\begin{tabular}{@{}ll@{}}
\toprule
\textbf{Hyperparameter} & \textbf{Value} \\
\midrule
Temperature & 0.7 \\
Max New Tokens & 2000 \\
API Version & 11-20 \\
\bottomrule
\end{tabular}
\end{table}

\label{app:hyperparams}

\begin{figure}[h!]
    \centering
    \makebox[\textwidth][c]{\includegraphics[width=1.3\textwidth]{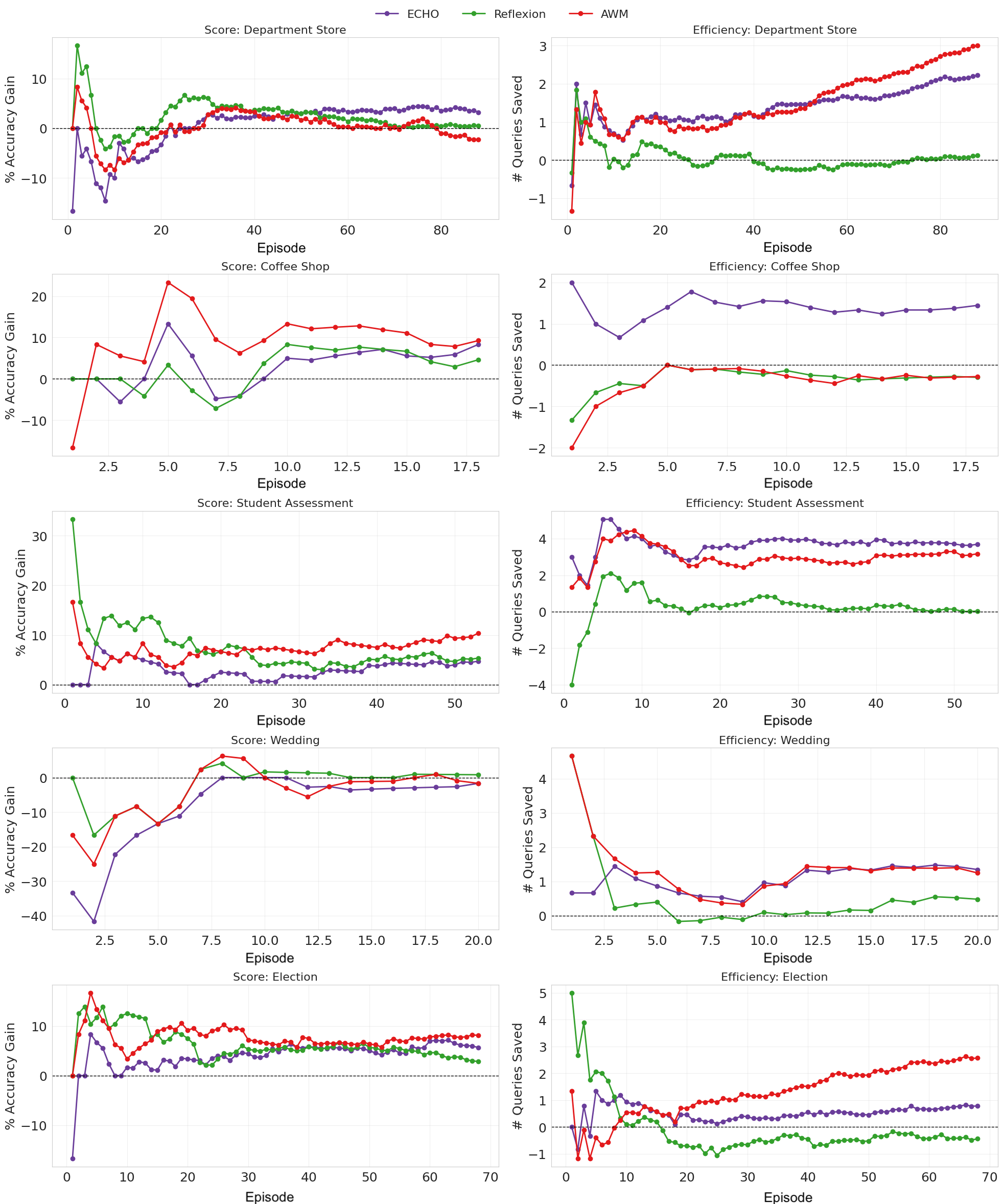}}
    \caption{We chose 5 organizations from PeopleJoinQA, maximizing variation amongst number of people in the organization and total number of queries. No offline method consistently outperforms the baseline on both accuracy and efficiency for all organizations.}
    \label{fig:peoplejoinqa-breakdown}
\end{figure}

\FloatBarrier

\end{document}